\documentclass[11pt]{article}
\usepackage[letterpaper,margin=1in]{geometry}
\usepackage[T1]{fontenc}
\usepackage{times}
\usepackage{microtype}
\usepackage{amsmath,booktabs,array}
\IfFileExists{amsthm.sty}{\usepackage{amsthm}}{%
  \newenvironment{proof}{\par\noindent\textit{Proof.}\ \ignorespaces}{%
    \unskip\nobreak\hfill\rule{.7ex}{.7ex}\par}%
}
\usepackage{graphicx}
\usepackage{eso-pic}
\usepackage[round,authoryear]{natbib}
\usepackage[hidelinks]{hyperref}

\makeatletter
\renewcommand\section{\@startsection{section}{1}{\z@}%
  {-2.6ex plus -.7ex minus -.2ex}{1.45ex plus .15ex}{\normalfont\Large\bfseries}}
\renewcommand\subsection{\@startsection{subsection}{2}{\z@}%
  {-2.1ex plus -.55ex minus -.2ex}{1.0ex plus .15ex}{\normalfont\large\bfseries}}
\makeatother

\DeclareRobustCommand{\textsc}[1]{\leavevmode\ArchiveSmallCaps#1\relax}
\def\ArchiveSmallCaps#1#2\relax{#1\scalebox{.79}[.77]{\MakeUppercase{#2}}}
\newcommand{\None}{\textsc{None}}
\newcommand{\Scenario}{\textsc{Scenario}}
\newcommand{\Persona}{\textsc{Persona}}
\newtheorem{proposition}{Proposition}
\newcommand{\repository}{\mbox{\url{https://github.com/rachael1216/fragile-ties}}}
\newcommand{\epigraphline}[2]{\begin{quote}\small\itshape #1\par\raggedleft\normalfont #2\end{quote}}

\title{The Tie That Wasn't\\[0.3em]
\large How Weighting and Sample Composition Flip Rankings of Supportive Replies}
\author{\normalsize Ruiqi Liu$^{1,2,\dagger}$, Yanlan Kang$^{1,2,\dagger}$, Caoyi Shen$^1$, and Tianchen Yang$^{3,\ast}$\\[0.5em]
\small $^1$Fudan University, 220 Handan Road, Yangpu District, Shanghai 200433, China\\
\small $^2$Institute for Medical Philosophy and Future Artificial Intelligence, Baotou 014010, Inner Mongolia, China\\
\small $^3$School of Psychology and Institute of Chinese Cultural Psychology,\\
\small Shanghai Jiao Tong University, 1954 Huashan Road, Xuhui District, Shanghai 200030, China\\
\small $^\dagger$Joint first authors. \quad $^\ast$Correspondence: \texttt{tianchenyang@sjtu.edu.cn}\thanks{This work was supported by the Youth Project of the National Social Science Fund of China in Education (CBA260278).}}
\date{}
\begin{document}
\AddToShipoutPictureFG*{%
  \AtPageUpperLeft{\raisebox{-73bp}[0pt][0pt]{%
    \makebox[\paperwidth][c]{%
      \raisebox{-.5\height}{\includegraphics[height=42bp]{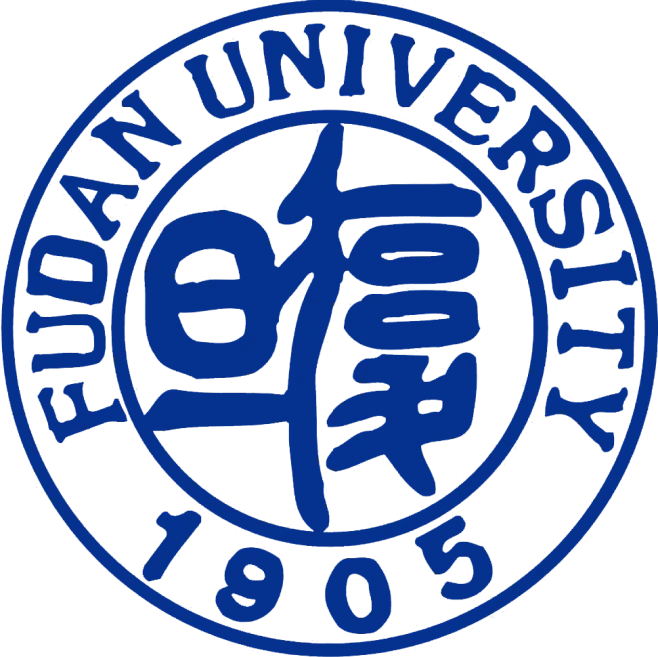}}%
      \hspace{12bp}%
      \raisebox{-.5\height}{\includegraphics[width=226bp]{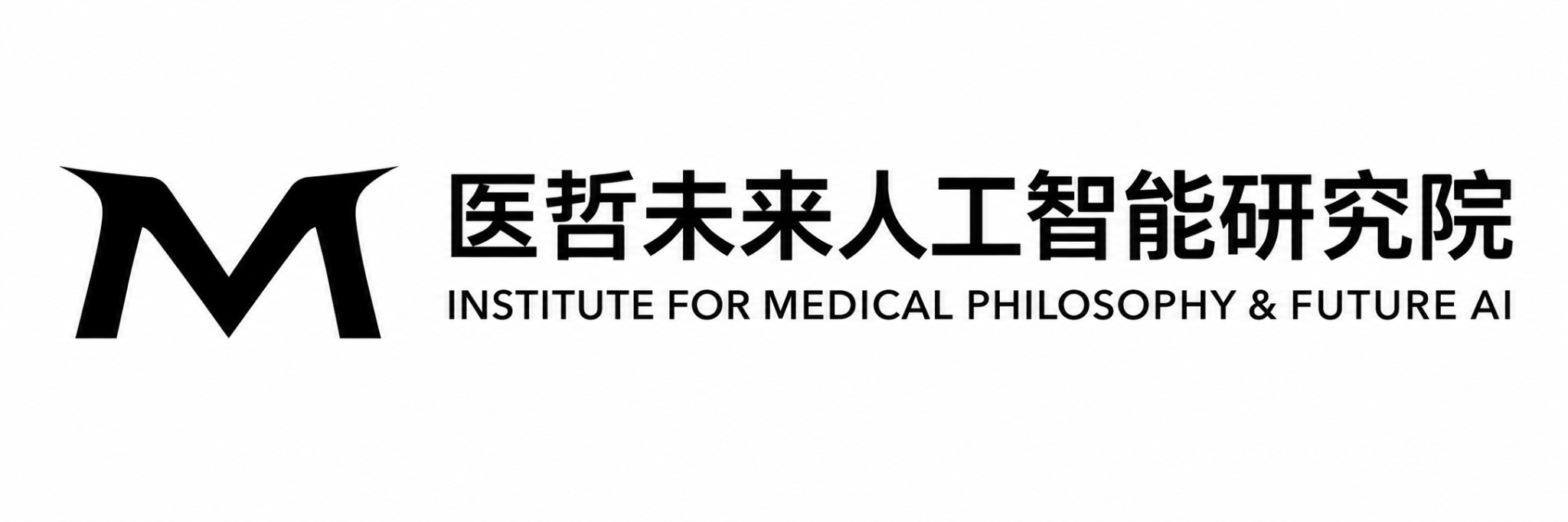}}}}}}
\maketitle

\begin{abstract}
A near tie on a measure of supportive dialogue can conceal sizable differences in the acts for which the measure gives credit. In an archive of 240 replies from 219 simulated English and Chinese dialogues, standardization to a common language and user-style mix gives \Scenario{} 7.09 percentage points less validation than the baseline and 6.55 points more offers of future contact, a difference of $-0.27$ points in the equal-weight mean. We develop a secondary audit in four parts. Direct standardization traces the passage from the original group compositions to a common target. Exact, two-parameter crossover analysis locates every change of leader and gives a deterministic criterion for the stability of its margin. Enumeration inverts the archived projection of the eleven binary codes, and an allocation-invariant identity recovers a weighted within-group offer gap of 23.65 percentage points. Sharp bounds then give the attainable extremes of contrasts at a common model mix and in specified subsets of turns. In the two-act comparison, standardization takes \Persona{} from last to first at equal weights, with a margin of 0.092 points. Its maximal lead is 0.186 points and occurs where the other two groups tie. The estimands, the conditions that identify them, and the corresponding archival inputs are set out together.
\end{abstract}

\section{Introduction}
\epigraphline{Far better an approximate answer to the right question, which is often vague, than an exact answer to the wrong question.}{John W. Tukey \citeyearpar[p.~13]{tukey1962}}

\begin{figure}[!htbp]
\centering
\includegraphics[width=.54\linewidth]{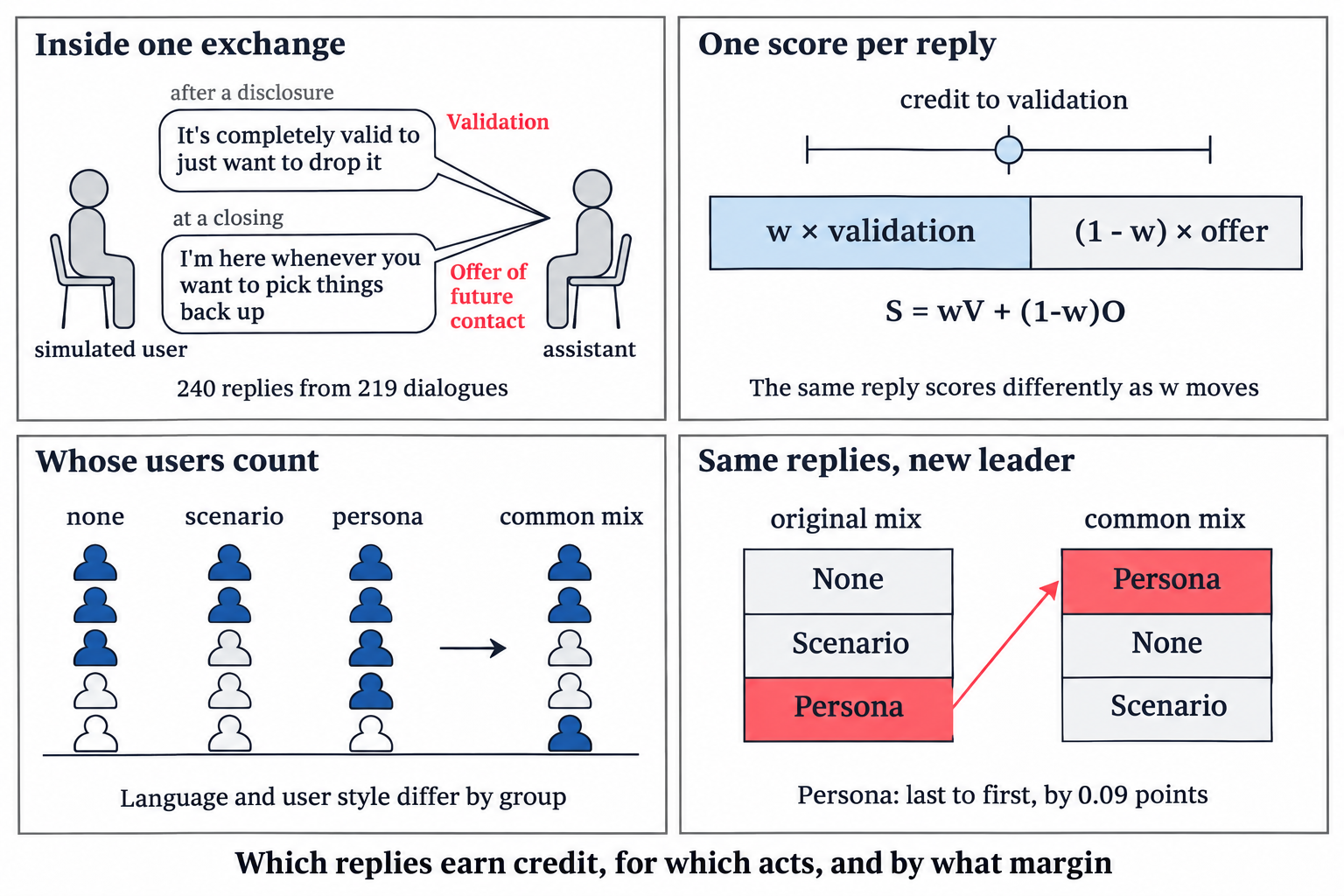}
\caption{Two acts in simulated dialogue and the choices by which an evaluation gives them credit. At a common composition, the two contrasts of \Scenario{} with \None{} are $-7.09$ and $+6.55$ percentage points, with an equal-weight contrast of $-0.27$ points. At those weights, standardization takes \Persona{} from third place to first.}
\label{fig:teaser}
\end{figure}

A simulated user wanted to drop a subject, and the assistant acknowledged the feeling. Another user said that they would return if they needed anything, and the assistant offered to take up the exchange at a later time. In Figure~\ref{fig:teaser}, one act attends to what has been disclosed and the other makes a subsequent exchange available. An average permits a rise in one to compensate for a fall in the other.

A further decision concerns whose conversations enter the average. A prompt group with more anxious turns or more turns in one language is evaluated on a different distribution of occasions from its rivals. Upon fixing a common distribution, the evaluator can compare how the groups respond to the same mixture of occasions. Upon fixing the weights, the evaluator can say which acts are rewarded and how much each can compensate for another. Both decisions have numerical consequences even when the replies and their annotations are held fixed.

The present audit asks what a reader can reconstruct from a release of counts, rates, and projected profiles.\footnote{The companion code and retained audit inputs are indexed in Appendix~\ref{app:repro} and at \repository.} Its first part separates the score from the distribution over which its components are averaged. We compare observed and directly standardized rates, and trace the whole path between the two. Its second part classifies the exact changes of rank over the joint choice of composition and weights. We record the extent of a leadership region and the largest margin attainable therein, and give a separate criterion for stability under perturbations to reported rates. Its third part inverts the retained projection and identifies a group-adjusted association from the pooled overlap and group margins. Its fourth part characterizes the extrema consistent with recorded counts, for comparisons that require a finer cross-classification than the archive supplies.

A composite near tie, a change of leader, and the identification of an association are statements about different functions of the same archive. The audit places these functions alongside the records from which each can be recovered.

\section{Responsiveness, Availability, and the Construction of Scores}
\label{sec:literature}
\subsection{The acts and their occasions}
In the attachment literature, a partner acts as a safe haven insofar as the partner is both accessible and responsive to immediate need \citep{collins2000}. What is received as responsive support also depends upon the recipient's interpretation of the act \citep{maisel2009,reis2015}. Acknowledgment gives the speaker reason to believe that the feeling has been understood. An offer of future contact makes provision for a later occasion, which can be a useful aspect of continuity across repeated encounters \citep{bickmore2005}. These acts can appear together, and each can occur separately. Recording both acts allows us to see how a score gives them credit.

The prospect of continued contact has a particular place in the organization of an exchange. Conversation analysis distinguishes the work participants do to close an encounter from the work they do within the current topic \citep{schegloff1973}. The offer in Figure~\ref{fig:teaser} can acknowledge the user's wish to take leave and keep open a subsequent exchange. In another position, similar words can carry a different implication about the relation between the user and the assistant. Language that invites continuing interaction is among the cues examined in relational-boundary audits \citep{demeocq2026}. The archived code \texttt{invite\_dependency} records the offer itself. The evaluative credit given to that offer is a separate choice.

Studies of emotional support conversation distinguish strategies and examine their place in the encounter \citep{liu2021,kang2024}. In human peer support, model assistance has been studied as a means of improving the production of empathic replies \citep{sharma2023}. The unit here is the annotated act in a generated reply. A coder can ask whether the wording acknowledges a feeling, invites disclosure, or offers continued contact. A benchmark designer can then state whether and to what extent the score should reward that wording. This order matters when an offer of later availability and a response to immediate feeling serve different ends.

\subsection{A rank as a comparison at a specified composition}
The usefulness of a ranking is a function of the purposes of those who consult it \citep{ethayarajh2020}. Its value depends as well on how test outcomes are paired \citep{peyrard2021}, what items enter the comparison \citep{kovatchev2024}, and how a judgment is elicited \citep{rottger2024}. Each design fixes a comparison that can be varied in thought or in calculation. The archive allows us to vary the weights and the composition while holding the scored replies fixed.

The statistical point of departure is direct standardization \citep{kitagawa1955}. An aggregate rate has a within-cell component and a composition, and a comparison at common shares sets apart the effect of averaging over different occasions. We use a continuous path to extend the question from two selected compositions to the arrangements between them. The corresponding geometric object is the set of weights and compositions at which a group leads.

Some comparisons require a joint table in addition to the standardized rates. Partial identification \citep{manski2003} treats the observed margins as constraints on that table. The feasible allocations then supply attainable extrema. This yields a common discipline for the whole audit. Retain the reported totals, state the function of the table in question, and determine its value or its extrema over the admissible tables.

\section{The Archive and Its Information Structure}
\label{sec:archive}
\subsection{Sampling, grouping, and coverage}
The archive contains 240 replies from 219 conversations, 21 of which contribute two replies apiece. English and Chinese each contribute 120 replies. The simulated user styles are anxious, avoidant, and secure, with respective counts of 148, 58, and 34. The three prompt regimes are the baseline \None{} (69 replies), situational framing \Scenario{} (78), and relational-style persona framing \Persona{} (93).

\begin{figure}[!htbp]
\centering
\includegraphics[width=77mm]{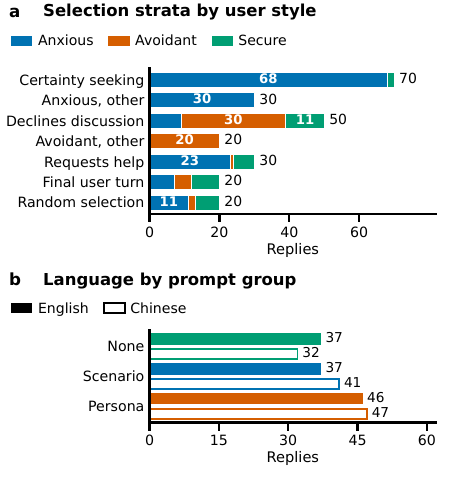}
\caption{Sample composition. Panel (a) gives counts within the seven selection strata by simulated user style. Panel (b) gives the English and Chinese counts by prompt regime.}
\label{fig:sample}
\end{figure}

A coding packet included the selected user turn and assistant reply, up to three preceding exchanges, and the next user turn where one existed. Four batches of 60 replies enriched particular contexts. The selection strata comprised 70 turns seeking certainty, 30 other anxious turns, 50 turns declining discussion, 20 other avoidant turns, 30 seeking help, 20 final user turns, and 20 turns selected at random. Fifty-nine packets had no following turn. Selection as a final user turn and the presence of a following turn are recorded at different levels. Figure~\ref{fig:sample} displays the coverage.

\begin{table}[!htbp]
\centering
\begin{tabular}{llrrrr}
\toprule
Language & Simulated user style & \None{} & \Scenario{} & \Persona{} & Total\\
\midrule
English & Anxious & 27 & 22 & 26 & 75\\
& Avoidant & 6 & 10 & 15 & 31\\
& Secure & 4 & 5 & 5 & 14\\
Chinese & Anxious & 21 & 22 & 30 & 73\\
& Avoidant & 7 & 11 & 9 & 27\\
& Secure & 4 & 8 & 8 & 20\\
\midrule
& Total & 69 & 78 & 93 & 240\\
\bottomrule
\end{tabular}
\caption{The complete prompt regime by language by simulated user style cross-tabulation. The last column supplies the numerators of the common shares.}
\label{tab:cross-style}
\end{table}

\begin{table}[!htbp]
\centering
\begin{tabular}{@{}lp{.63\linewidth}@{}}
\toprule
Symbol & Meaning\\
\midrule
$g,k$ and $h$ & Prompt groups and a language by style cell\\
$w$ and $\lambda$ & Weight on validation and fraction of the common mix\\
$R_g^{(b)}$, $S_g$, $L_g$ & Rate for code $b$, composite score, and margin over the strongest rival\\
$\mathcal I$ & Retained information for the audit\\
\bottomrule
\end{tabular}
\caption{Notation for the scoring and composition arguments.}
\label{tab:notation}
\end{table}

\begin{table}[!htbp]
\centering
\begin{tabular}{llrrrr}
\toprule
Assistant & User simulator & \None{} & \Scenario{} & \Persona{} & Total\\
\midrule
\texttt{gem31lite} & \texttt{dsv4flash} & 10 & 10 & 20 & 40\\
& \texttt{gpt54mini} & 17 & 18 & 15 & 50\\
\texttt{haiku45} & \texttt{dsv4flash} & 8 & 11 & 11 & 30\\
& \texttt{gpt54mini} & 15 & 8 & 14 & 37\\
\texttt{qwen35plus} & \texttt{dsv4flash} & 10 & 14 & 15 & 39\\
& \texttt{gpt54mini} & 9 & 17 & 18 & 44\\
\midrule
& Total & 69 & 78 & 93 & 240\\
\bottomrule
\end{tabular}
\caption{Counts by prompt group and the archived aliases of the assistant and user simulator. Appendix~\ref{app:bounds} gives the finer common cells used to bound a standardized contrast.}
\label{tab:cross-model}
\end{table}

Table~\ref{tab:cross-style} shows the six cells used for direct standardization. Each group occupies all six, and the cell sizes range from 4 to 30. The model role counts in Table~\ref{tab:cross-model} yield assistant totals of 90, 67, and 83, and simulator totals of 109 and 131, respectively, in the order in which the aliases occur. Model comparisons later use the intersection of occupied assistant by simulator by language by style cells.

\subsection{Coding and agreement}
The eleven binary fields distinguish a guarantee, affirmation of worth, validation of feelings, acknowledgment of uncertainty, invitation to disclose, bare shift of topic, return of choice, referral to human support, invitation to future contact, AI intimacy, and a template response. In functional terms, the fields cover acknowledgment and elaboration, agency and referral, present relational language and future availability, and features of assurance and form. A reply may receive several codes simultaneously. The focal comparison gives credit to validation and to the offer of future contact. Appendix~\ref{app:codebook} gives the field definitions, original rates, standardized rates, and arithmetic contrasts for the entire profile.

An automated judge supplied the archived decisions. The question of how an automated judge agrees with a reference is distinct from the question of how two human coders agree about a construct \citep{zheng2023}. The human summary reports three coders and a sample of 20, with $\kappa=.8674$ among the coders and $\kappa=.9453$ for human--model agreement. A separate comparison of three model-generated annotation sheets with the archived judge gives disagreement rates of 17.9\% for validation and 6.7\% for future contact. The first comparison supplies aggregate agreement coefficients. The second measures disagreement separately for the two fields.

\subsection{The retained information}
Write the information supplied to the audit as
\[
\mathcal I=\{\mathcal N,\mathcal R,\mathcal P,\mathcal T,\mathcal A\}.
\]
Here $\mathcal N$ consists of prompt, language, style, and model-cell counts. $\mathcal R$ contains the original and standardized code rates and group totals, and $\mathcal P$ contains the ordered code means, loadings, projected coordinates, and frequencies derived from the retained profile summaries. $\mathcal T$ contains the turn subset sizes and pooled, template-stratified contingency summaries. $\mathcal A$ contains the aggregate agreement information. By working separately with each level of aggregation, we can specify which components of $\mathcal I$ identify a quantity. For instance, the standardized ranking uses $\mathcal R$, the pooled focal table can be recovered from $\mathcal P$, and a logical interval for a subset uses its size from $\mathcal T$ together with that pooled table.

\section{Five Results for a Secondary Audit}
\label{sec:theory}
\subsection{Direct standardization and the composition path}
Let $g$ and $k$ denote prompt groups, $h$ one of the six language by style cells, and $b$ a binary code. Write $r_{gh}^{(b)}=100m_{gh}^{(b)}/n_{gh}$ for the cell rate, $q_{gh}=n_{gh}/n_g$ for the group share and $q_h=\sum_g n_{gh}/240$ for the common share. A group rate along the path has the form
\begin{equation}
R_g^{(b)}(\lambda)=\sum_h q_{gh}(\lambda)r_{gh}^{(b)},\qquad
q_{gh}(\lambda)=(1-\lambda)q_{gh}+\lambda q_h,\quad 0\le \lambda\le 1.
\label{eq:mix}
\end{equation}
Thus $R_g^{(b)}(\lambda)=(1-\lambda)R_g^{(b)}(0)+\lambda R_g^{(b)}(1)$. The path uses the two retained rates at its endpoints. For a focal contrast $g-k$, the following decomposition gives an exact accounting at the target:
\begin{align}
R_g^{(b)}(0)-R_k^{(b)}(0)
&=\sum_h q_h\bigl(r_{gh}^{(b)}-r_{kh}^{(b)}\bigr)\nonumber\\
&\quad+\sum_h\bigl((q_{gh}-q_h)r_{gh}^{(b)}
 -(q_{kh}-q_h)r_{kh}^{(b)}\bigr).
\label{eq:decomposition}
\end{align}
The first term compares cell rates at common weights. The second gives the change due to using each group's own shares. Since the path changes only the shares, a difference between contrasts at two values of $\lambda$ has an interpretation fixed by the decomposition.

\subsection{Crossover geometry and the size of a lead}
Let $V_g(\lambda)$ be the validation rate and $O_g(\lambda)$ the future-contact rate, both on the 0--100 scale. The score and its margin are
\begin{equation}
S_g(w,\lambda)=wV_g(\lambda)+(1-w)O_g(\lambda),\qquad
L_g(w,\lambda)=S_g(w,\lambda)-\max_{k\ne g}S_k(w,\lambda).
\label{eq:score}
\end{equation}
Here $0\le w\le1$ as in Table~\ref{tab:notation}. A positive $L_g$ denotes strict leadership and a zero margin denotes a tie for the lead.

\begin{proposition}[Weight-axis crossings and extremal margins]
\label{prop:weight}
At a fixed composition the score contrast of two groups is affine in $w$. Opposite signs of the endpoint contrasts give a unique interior crossing. Sorting the crossings partitions the weight axis into intervals of constant order, and the maximal signed margin of any group is attained at an endpoint of the axis or at a crossing.
\end{proposition}

\begin{proof}
At fixed $\lambda$ write $\delta_O=O_g-O_k$ and $\delta_V=V_g-V_k$. The contrast is $\delta_O+w(\delta_V-\delta_O)$. If the slope is nonzero, its zero is
\begin{equation}
 w_{gk}^{*}=\frac{\delta_O}{\delta_O-\delta_V}.
\label{eq:cross}
\end{equation}
An interior zero has opposite-signed endpoint contrasts, and the converse follows by solving the displayed equation. Equal endpoint contrasts yield parallel score curves or curves that agree throughout. On each open interval between distinct crossings all nonidentical pairs have constant order. There $L_g$ equals the difference between two fixed affine scores, hence achieves its largest value on the closure at one of its endpoints. Taking the greatest value over the closures gives the stated result.
\end{proof}

\begin{proposition}[The two-parameter leader partition]
\label{prop:joint}
Over the square $(w,\lambda)\in[0,1]^2$, each pairwise contrast is affine in either argument with the other held fixed. For the archived focal rates, \Persona{} has a strict lead precisely between its two rival crossings, in the order specified below. The onset of its leadership is the unique zero in $[0,1]$ of a quadratic numerator.
\end{proposition}

\begin{proof}
Each component rate is affine in $\lambda$. Substitution into $S_g$ permits a constant, a term in each argument separately, and a product term. The group $g$ leads strictly on the intersection of $S_g-S_k>0$ over its rivals. For fixed $\lambda$ their boundaries are given by Eq.~\eqref{eq:cross}. Thus each interior pairwise boundary is a ratio of affine functions of $\lambda$.

For the three archived groups let $d_b$ denote \Scenario{} minus \None{} on code $b$, and $p_b$ denote \Persona{} minus \None{}. The endpoint rates put \Persona{} between the rivals on each act, in reverse order across the acts. Interpolation preserves both orders. Define
\[
A=(p_V-d_V)-(p_O-d_O),\qquad B=p_O-p_V.
\]
Both slopes are positive. The two score advantages of \Persona{} are $A(w-\ell)$ over \Scenario{} and $B(u-w)$ over \None{}, where $\ell=(d_O-p_O)/A$ and $u=p_O/B$. Both are positive exactly for $\ell<w<u$. The maximum of their minimum occurs at the intersection of the two lines. At this point \Scenario{} and \None{} tie, and
\begin{align}
w^*&=\frac{A\ell+Bu}{A+B}=\frac{d_O}{d_O-d_V},\label{eq:lead-cross}\\
L_{\text{\Persona}}(w^*,\lambda)&=\frac{AB(u-\ell)}{A+B}
=\frac{d_O p_V-d_V p_O}{d_O-d_V}.
\label{eq:persona-peak}
\end{align}
The denominator is positive throughout. The numerator is quadratic, and its two roots calculated from the archived endpoint rates are exhibited in Appendix~\ref{app:crossing}. Exactly one belongs to the unit interval. Since the maximal signed margin is negative at the original mix and positive at the common mix, the sign changes from negative to positive at that root. The two boundaries coincide there, and beyond it the strict interval of leadership opens.
\end{proof}

Equation~\eqref{eq:persona-peak} distinguishes the size of the winning region from the margin available within it. The first may be read in units of analyst discretion over $w$. The second is read in score points. Their relation involves $A$ and $B$, so an interval of appreciable width can correspond to a small margin when its two bounding score contrasts have shallow slopes. This distinction bears directly on the interpretation of the archive.

\subsection{Group-adjusted aggregation from an unallocated overlap}
Let $n_g$ be the number of replies in group $g$, $v_g$ the validation count, $o_g$ the offer count, and $j_g$ the count with both. A difference in offer proportions within $g$, taken as without validation minus with validation, is
\[
D_g=\frac{o_g-j_g}{n_g-v_g}-\frac{j_g}{v_g}.
\]
Throughout the archived groups, $0<v_g<n_g$. Set $u_g=v_g(n_g-v_g)/n_g$.

\begin{proposition}[An invariant within-group contrast]
\label{prop:pooled}
The weighted mean $\sum_g u_gD_g/\sum_g u_g$ follows from the group margins and the total overlap $J=\sum_g j_g$. Each group's sharp extrema follow by intersecting its own feasible interval with the pooled sum constraint.
\end{proposition}

\begin{proof}
Multiplication by $u_g$ gives the identity
\begin{equation}
u_gD_g=\frac{v_go_g}{n_g}-j_g.
\label{eq:within}
\end{equation}
Summation fixes the numerator at $\sum_g v_go_g/n_g-J$. For each group the range of $j_g$ before imposing $J$ is the integer interval with lower limit $a_g=\max(0,v_g+o_g-n_g)$ and upper limit $b_g=\min(v_g,o_g)$. The other groups together can attain every integer between $\sum_{k\ne g}a_k$ and $\sum_{k\ne g}b_k$, since sums of integer intervals are integer intervals. Thus the sharp feasible endpoints of $j_g$ are
\begin{align*}
j_g^{-}&=\max\left(a_g,J-\sum_{k\ne g}b_k\right),&
j_g^{+}&=\min\left(b_g,J-\sum_{k\ne g}a_k\right).
\end{align*}
For any integer in this interval the complement can be assigned within the corresponding intervals, giving a nonnegative table in every group. Since $D_g$ decreases in $j_g$, reversing these endpoints gives its sharp extrema.
\end{proof}

The weights have a familiar regression interpretation. With $x_i$ and $y_i$ the validation and offer indicators, respectively, subtracting their group means gives a within-group sum of products $j_g-v_go_g/n_g$ and a sum of squared centered values of $x_i$ equal to $u_g$. The coefficient on $x_i$ in the regression of $y_i$ on $x_i$ and group intercepts is consequently $-\sum_g u_gD_g/\sum_g u_g$. The group margins and the pooled overlap supply those same sums. Thus a group-adjusted association can be calculated from a release at this level of detail, as can the separate group bounds in Appendix~\ref{app:projection}.

\subsection{Inverting the archived projection}
For $z$ in the set $\{0,1\}^{11}$ let
\begin{equation*}
 x(z)=L^\top(z-\mu)/\sqrt{11},
\tag{P}\label{eq:project}
\end{equation*}
where $L$ contains the two archived loading vectors and $\mu$ is the pooled mean of the eleven binary codes, reconstructed from their original group rates and counts. The frequencies of the archived coordinate pairs describe a distribution over this finite set of possibilities.

\begin{proposition}[Finite projection inversion]
\label{prop:inverse}
For the retained loading coefficients the map in Eq.~\eqref{eq:project} is injective on the $2^{11}$ binary vectors. Every archived point has one candidate within the matching tolerance of $10^{-10}$. The pooled joint table is therefore uniquely recovered by assigning each projected frequency to the corresponding binary vector and zero to the other vectors.
\end{proposition}

\begin{proof}
Enumerate the candidate vectors in the archive's code order. For distinct $z,z'$ the distance between their projections is $\|L^\top(z-z')\|_2/\sqrt{11}$, which is independent of the common center. Appendix~\ref{app:projection} gives the finite calculation of its minimum, approximately $7.1386\times10^{-4}$. It is positive, and exceeds twice the tolerance by several orders of magnitude. A point within tolerance of two candidates would place the candidates at most twice the tolerance apart by the triangle inequality. Direct comparison of the retained coordinates with the complete list gives a candidate within tolerance for each. Each point therefore has exactly one match. Assigning counts by these matches gives 83 occupied profiles of total frequency 240. Summation over coordinates reproduces the retained one-code totals, and summation over the two focal coordinates gives the joint counts $(43,30,139,28)$ in the order neither, offer alone, validation alone, and both.
\end{proof}

Inverting the focal projection is a question about separation on the archived alphabet. The recovered margins provide a check before those counts enter any association or bounding calculation.

\subsection{Deterministic stability of reported margins}
A score margin and the width of its leadership interval describe different perturbations. The width concerns changes of the weight or composition. A margin also gives a direct criterion for how much the reported rates can change while preserving the same ordering at a fixed weight and composition.

\begin{proposition}[Perturbation bounds for rates and scores]
\label{prop:margin}
If every reported score may change by at most $\epsilon$ points, a strict lead of size $m>0$ is preserved under every such change exactly when $m>2\epsilon$. At fixed $w$, if each group's validation and offer rates may separately change by at most $\epsilon_V$ and $\epsilon_O$, the corresponding threshold is
\[
m>2\{w\epsilon_V+(1-w)\epsilon_O\}.
\]
The latter threshold is exact when the changes in the two directions attaining equality are admissible for the leader and the nearest rival.
\end{proposition}

\begin{proof}
A score contrast can decrease by at most the sum of the admissible downward change for the leader and the upward change for its rival. For a common score allowance $\epsilon$ those changes sum to $2\epsilon$. The nearest rival and the leader can attain the two extremes together under the stated independent score allowances. Equality yields a tie, and strict excess of the combined allowance over $m$ permits a reversal. With component-specific allowances the weighted score change has magnitude at most $w\epsilon_V+(1-w)\epsilon_O$. Simultaneous changes of this magnitude in opposite directions give the same argument, provided that the component rates remain within their domains. When such domains truncate an extreme, the sum of the attainable downward and upward allowances provides the corresponding pairwise criterion.
\end{proof}

\section{Results for the Composite and the Full Profile}
\label{sec:results}
\subsection{The joint occurrence of the focal acts}
\begin{figure}[!htbp]
\centering
\begin{minipage}[t]{.475\linewidth}
\vspace{0pt}\centering
\includegraphics[width=\linewidth]{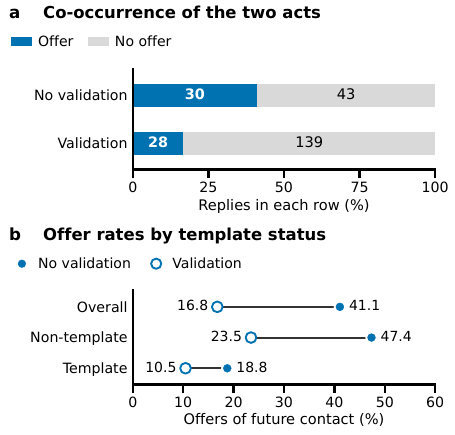}
\caption{Co-occurrence of validation and offers of future contact. Panel (a) gives the counts for offers and their complements with and without validation. Panel (b) gives the offer rates in the full set and within the two levels of the template code.}
\label{fig:cooccur}
\end{minipage}\hfill
\begin{minipage}[t]{.475\linewidth}
\vspace{0pt}\centering
\includegraphics[width=\linewidth]{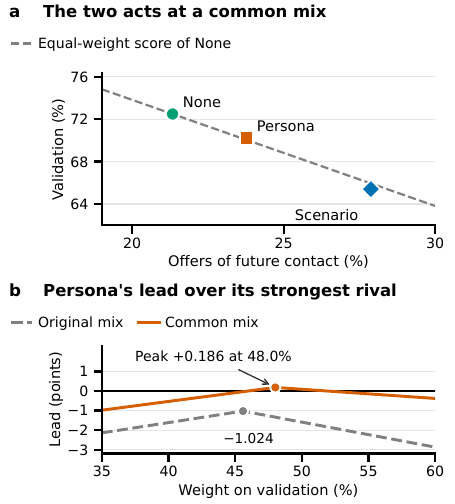}
\caption{Focal rates and the margin of \Persona{} over its stronger rival. Panel (a) places the common-composition rates in the two-act plane. The dashed line consists of pairs with the equal-weight score of \None{}. Panel (b) compares the signed margin at the original and common compositions and locates its common-composition maximum.}
\label{fig:ranks}
\end{minipage}
\end{figure}

The recovered table puts offers of future contact in 30 of the 73 replies without validation (41.1\%) and in 28 of the 167 with validation (16.8\%). The pooled difference is 24.3 percentage points. The same ordering occurs among replies without the template code (47.4\% versus 23.5\%) and among those with it (18.8\% versus 10.5\%). Figure~\ref{fig:cooccur} presents the joint frequencies and the conditional comparisons together.

Under within-group independence, the number of joint positives expected from the retained margins is $\sum_g v_go_g/n_g=39.8689$. With 28 joint positives and a weight sum of 50.1956, Proposition~\ref{prop:pooled} gives a within-group gap of 23.65 percentage points. The gap has the same sign and nearly the same size as the pooled contrast. Here the persistence of the difference across these two calculations can be established without allocating the overlap to individual groups. The group-specific feasible intervals in Appendix~\ref{app:projection} give those allocations their separate interpretation.

\subsection{A near tie and a change of leader}
At the common composition, greater future availability accompanies less validation in the comparison of \Scenario{} with \None{}. The original contrast in validation is $-12.71$ points, and its standardized counterpart is $-7.09$. For future contact the corresponding differences are $+10.65$ and $+6.55$ points. Accordingly, the two composition terms in Eq.~\eqref{eq:decomposition} are about $-5.62$ and $+4.10$ points respectively. Standardization leaves both directions intact and brings both act rates closer together. At equal weights the resulting contrast is $-0.27$ points, with 0.36 points separating the highest and lowest of the three scores in Table~\ref{tab:focal}.

\begin{table}[!htbp]
\centering
\begin{tabular}{lrrrrrr}
\toprule
& \multicolumn{3}{c}{Original composition} & \multicolumn{3}{c}{Common composition}\\
\cmidrule(lr){2-4}\cmidrule(l){5-7}
Group & Validation & Offer & Score & Validation & Offer & Score\\
\midrule
\None{} & 76.81 & 18.84 & 47.83 & 72.49 & 21.33 & 46.91\\
\Scenario{} & 64.10 & 29.49 & 46.79 & 65.40 & 27.88 & 46.64\\
\Persona{} & 68.82 & 23.66 & 46.24 & 70.24 & 23.77 & 47.00\\
\bottomrule
\end{tabular}
\caption{Original and standardized focal rates (percent) with their equal-weight mean scores. All differences and crossings use the unrounded rates.}
\label{tab:focal}
\end{table}

Figure~\ref{fig:ranks}(a) places each common-composition result in the plane of its two components. The dashed line through \None{} is the set of pairs with its equal-weight score. A point can move a considerable distance in that plane while remaining close to the line. The remaining points are just such cases. An equal-weight rank answers a question about position relative to that line, whereas the two coordinates give a direct account of the acts that produced the rank.

At the original composition \Scenario{} leads before $w=.4558$ and \None{} leads thereafter, with a tie at the boundary. The maximum signed margin of \Persona{} is $-1.02$ points. Upon standardization, the \Persona{} crossings occur in the opposite order. From 45.9\% to 52.0\% of the score assigned to validation, it leads both other groups, with ties at the endpoints. The widest margin is $+0.186$ points at $w=.4801$, where the two rivals tie. At $w=.5$ the margin is $+0.092$ points. The margin is therefore best reported alongside the weight and composition under which it arises.

\subsection{The phase diagram}
\begin{figure}[!htbp]
\centering
\includegraphics[width=.90\linewidth]{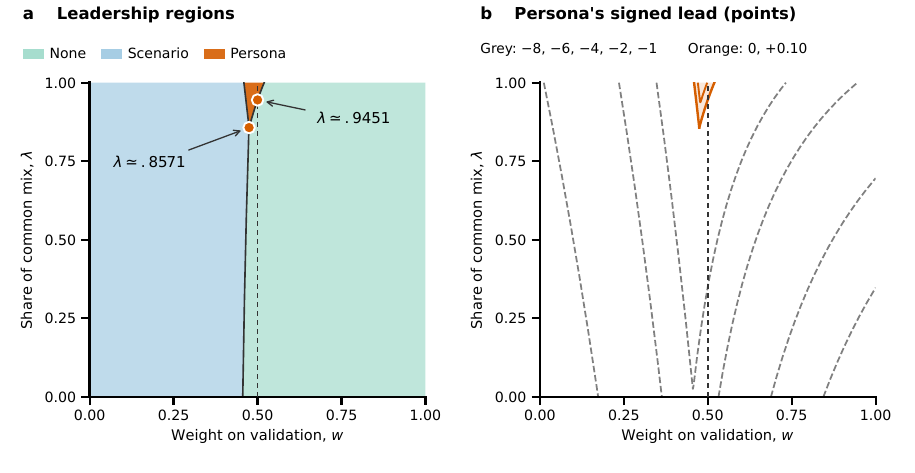}
\caption{Analytical phase diagram on the composition and weight square. Panel (a) identifies the leading prompt group, with equality on the separating boundaries. Panel (b) gives level sets of the signed \Persona{} margin. The onset composition is $\lambda\approx .8571$, while the equal-weight line meets the boundary at $\lambda\approx .9451$.}
\label{fig:phase}
\end{figure}

The two-parameter calculation puts the original and common rankings at the lower and upper edges of the unit square in Figure~\ref{fig:phase}. A vertical path through $w=.5$ describes what changing the composition does to the equal-weight rank. A horizontal line through any fixed $\lambda$ shows what changing the value placed on validation does to the ranking at that composition.

The first composition at which \Persona{} can share the lead is $\lambda\approx .8571$, the root in Proposition~\ref{prop:joint}. At this point the lower and upper boundaries coincide and all three groups tie. The equal-weight line enters the leadership region later, at $\lambda\approx .9451$. That second threshold comes from the linear equation $p_V+p_O=0$. At fixed $\lambda$ the best margin is reached where the two rival scores agree, which accounts for the meeting of two branches of the margin contours. The shape of the region records a joint condition on composition and credit, while its level sets measure the corresponding differences in the score.

\subsection{Eleven codes and the simplex of weights}
\begin{figure}[!htbp]
\centering
\includegraphics[width=77mm]{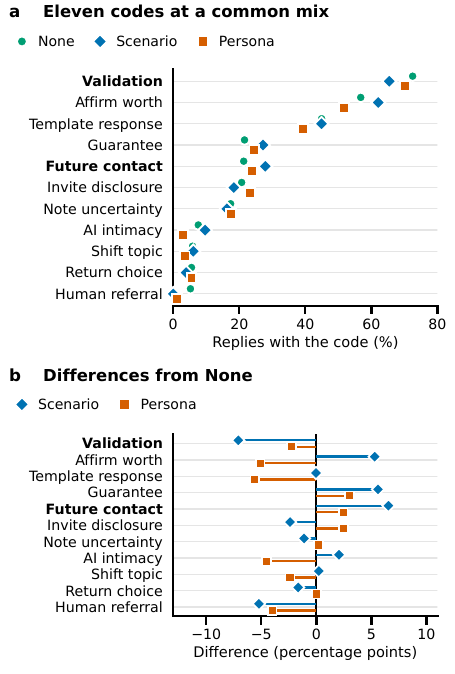}
\caption{Standardized profiles on all eleven codes. Panel (a) gives the rates sorted by the value for \None{}. Panel (b) gives each other group's difference from the baseline in percentage points.}
\label{fig:profiles}
\end{figure}

The fuller profile gives a direct account of why rankings depend on the acts credited. Whereas \Scenario{} validates feelings less often than \None{}, it affirms worth 5.30 points more often and supplies guarantees 5.60 points more often at the common composition. Its difference from \Persona{} is $+10.40$ points for affirming worth and $-4.87$ points for inviting disclosure. The latter contrast has the same sign as the contrast on validation. The two panels in Figure~\ref{fig:profiles} distinguish the absolute rates from their differences relative to the baseline.

The full range of nonnegative credit assignments is the simplex $\Delta_{10}=\{\alpha\ge0\mid\sum_{b=1}^{11}\alpha_b=1\}$. At the common composition let $\mathbf R_g$ denote the eleven rates of group $g$. A pairwise contrast is $\alpha^\top(\mathbf R_g-\mathbf R_k)$, hence ranges from the smallest to the largest of the eleven component contrasts. A group dominates another throughout the simplex when all its component contrasts have the requisite sign. The two focal acts span one edge of the simplex. All strict orderings on that edge are therefore available within the full simplex. They also extend to weights giving small positive credit to each remaining act, by continuity.

In the archived common-composition profiles, each pair takes both orders under some choice of code. \None{} leads on four simplex vertices, \Scenario{} on five, and \Persona{} on two. These are comparisons of code incidence under linear credit, with the chosen signs and weights specified as part of the scoring rule. For instance, giving all credit to reduced template use requires a different coding convention from giving credit to more template responses.

The maximum winning margin over the whole simplex is the solution of a linear program in $(\alpha,t)$, maximizing $t$ subject to $\alpha\in\Delta_{10}$ and
\[
 t\le\alpha^\top(\mathbf R_g-\mathbf R_k)\quad\text{for every }k\ne g.
\]
The solutions under the stated nonnegative incidence scoring are 4.29 points for \None{}, 6.14 for \Scenario{}, and 2.64 for \Persona{}. For \Persona{} the optimum lies on the edge joining guarantees and invitations to disclose. On the two-act edge, the same group can lead by at most 0.186 points. Appendix~\ref{app:codebook} gives the pairwise extrema and the linear-program solutions.

\section{Sharp Bounds from Retained Margins}
\label{sec:bounds}
\epigraphline{He who knows only his own side of the case, knows little of that.}{John Stuart Mill \citeyearpar{mill1859}}

\subsection{A common distribution of models, languages, and styles}
The common-composition comparison of Section~\ref{sec:results} uses language and style. We can instead ask what the same archive allows for comparisons at the same assistant and user-simulator composition as well. Let $j$ index cells defined by assistant, simulator, language, and style, with $h(j)$ the language by style cell. For $g\in\{\text{\None},\text{\Scenario}\}$ let $n_{gj}$ be the retained counts, let $C$ be the intersection of occupied cells, and let $r_j$ be cell $j$'s share of pooled replies restricted to $C$. There are 24 common cells in a union of 32. They cover 62 of the 69 replies in \None{} and 68 of the 78 in \Scenario{}. Table~\ref{tab:cross-model} gives the coarser model composition, and Appendix~\ref{app:bounds} lists every common cell.

For either focal behavior, let $x_{gj}$ count the positives assigned to cell $j$. If $M_g$ is the observed group positive count, and $R_g^{(b)}(1)$ is its standardized rate on the 0--100 scale, the integers must satisfy
\begin{align}
0\le x_{gj}\le n_{gj},\qquad
\sum_jx_{gj}&=M_g,\nonumber\\
\sum_j\frac{q_{h(j)}}{n_{g,h(j)}}x_{gj}&=\frac{R_g^{(b)}(1)}{100}.
\label{eq:model-constraints}
\end{align}
The constraints range over all occupied cells. The objective for a comparison at the common distribution of models is $100\sum_{j\in C}r_jx_{gj}/n_{gj}$. Retaining positive counts outside $C$ in the constraints allows them to account for the part of the marginal totals outside the comparison.

For exact integer scaling put $H_g=\operatorname{lcm}_h n_{gh}$ and $N_h=\sum_g n_{gh}$. Multiply the second equality by $240H_g$. Its coefficient in cell $j$ is then $N_{h(j)}(H_g/n_{g,h(j)})$. Integer right-hand sides are reconstructed at this scale from the retained rates and checked against their stored precision. The resulting integer linear programs give the least and greatest common-mix rates attainable in each arm.

At these constraints the two arm allocations are independent. If $(\ell_g,u_g)$ are the extrema for one arm, the least contrast is $\ell_{\text{\Scenario}}-u_{\text{\None}}$ and the greatest is $u_{\text{\Scenario}}-\ell_{\text{\None}}$. Each pair of endpoint allocations can coexist, since it satisfies constraints on disjoint sets of replies. Within each cell, any permitted integer $x_{gj}$ can be assigned to that number of its replies. Thus both endpoints are realizable under the retained marginal constraints for the act in question. The two bounds describe each act separately, with no assumption that the extrema across acts share a single joint allocation.

The computed intervals are $[-39.2,+18.0]$ points for validation and $[-20.3,+36.2]$ for future contact. The common-composition estimates that use language and style alone are included as reference points in Figure~\ref{fig:bounds}(a). The integer allocations in Appendix~\ref{app:bounds} exhibit the comparisons of opposite signs that the retained margins permit.

\begin{figure}[!htbp]
\centering
\includegraphics[width=77mm]{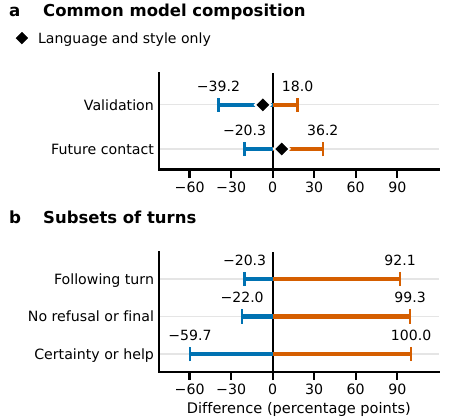}
\caption{Attainable bounds from retained margins. Panel (a) gives the two signed contrasts of \Scenario{} with \None{} at a common model distribution, with diamonds indicating standardization on language and style alone. Panel (b) gives contrasts of offer rates without and with validation in the specified turn subsets.}
\label{fig:bounds}
\end{figure}

\subsection{Turn-specific comparisons from subset sizes}
Suppose that the subset of interest has $m$ replies. For each potential allocation $(a,b,c,d)$ of profiles to that subset, in the order neither, offer alone, validation alone, and both, impose
\[
(a,b,c,d)\le(43,30,139,28),\qquad
a+b+c+d=m,\qquad a+b>0,\quad c+d>0
\]
componentwise, along with nonnegative integrality. Its offer gap in percentage points is $100\{b/(a+b)-d/(c+d)\}$. Every admissible allocation is attained by taking the specified number of replies from each pooled cell. The complement receives the residual counts. Enumerating these allocations therefore yields sharp extrema for the contrast under the stated subset information.

There are 181 replies with a following turn, and the compatible gap runs from $-20.3$ to $+92.1$ points. Excluding the refusal and final selection strata leaves 170 replies and an interval from $-22.0$ to $+99.3$. The certainty-seeking and help-request strata together have 100 replies, with bounds from $-59.7$ to $+100.0$. Figure~\ref{fig:bounds}(b) and Appendix~\ref{app:bounds} give these ranges and their extremal tuples. The three optimizations describe the information carried by each subset's size separately.

\section{What a Release Can Establish}
\label{sec:reporting}
\subsection{From records to claims}
An assertion that two groups performed alike on a composite concerns a specified average. The component rates reveal the consequences of that scoring rule, while the target distribution states whom the comparison represents. At that distribution, the lead interval gives the range over which the evaluator may alter the credit assigned to an act while preserving the leader.

Upon fixing both choices, the margin has a distinct use. The equal-weight lead of \Persona{} at the common composition is about 0.092 points. When reported scores may move independently, changes of 0.0462 points in opposite directions suffice to reverse its ordering with the nearest rival. By Proposition~\ref{prop:margin}, the loss of a strict lead first occurs at half the gap. That deterministic calculation gives a direct way to report a numerical tolerance beside the lead and its interval. The corresponding formula for separate tolerances on validation and contact can be stated in terms of the two components themselves.

The point comparisons and the bounds in Table~\ref{tab:information} use different parts of the retained information. The pooled overlap identifies the adjusted association, while a subset gap depends on which profiles the subset contains.

\begin{table}[!htbp]
\centering
\small
\begin{tabular}{@{}p{.28\linewidth}p{.38\linewidth}p{.26\linewidth}@{}}
\toprule
Retained evidence & Audit quantity & Identification\\
\midrule
Original and target code rates & Score at any $(w,\lambda)$ on the specified path and all crossover margins & Exact\\[0.25em]
Binary code order, two loadings, pooled means and projected frequencies & Pooled joint profile frequencies & Exact when the archive's finite candidate projections separate\\[0.25em]
Group focal margins and total joint positives & Weighted within-group gap and adjusted regression slope & Exact\\[0.25em]
The same margins and overlap & Individual group offer gaps & Sharp extrema from the feasible group overlaps\\[0.25em]
Model-cell sizes and group observed and language-by-style standardized margins & Each focal contrast at a common model mix & Sharp extrema from integer allocation\\[0.25em]
Pooled joint counts and one turn subset size & Offer gap in that subset & Sharp extrema from feasible subset counts\\[0.25em]
Dialogue and sampling links & Design-based uncertainty at the unit of selection & Requires a linked sample and a sampling design\\
\bottomrule
\end{tabular}
\caption{Information and identification in the secondary audit. Exact quantities use the tabulated rates or recovered counts as given. Sharp ranges record attainable extrema under the respective constraints.}
\label{tab:information}
\end{table}

\subsection{A standard for reporting comparisons}
A practical release begins with the component rates and definitions of the acts, the distribution of occasions over which the rates are compared, and the score weights. To this it can add the winning margin and the interval of weights, or the region in the joint plane, over which the same leader prevails. These quantities let a reader recompute the ranking and see where another choice of credit or comparison target changes the answer.

An association claim should name the retained cross-classification. A pooled joint count suffices for Proposition~\ref{prop:pooled}, and a subset size suffices for the corresponding sharp bounds. Links among disclosures, replies, and next turns would additionally allow a reader to examine whether an offer concludes or extends an exchange.

\section{Concluding Remarks}
A near tie in a social score can result from compensation between acts, and a first place can arise upon changing the distribution of occasions to which the score refers. Direct standardization supplies a common target and a continuous accounting of the change in composition. Exact crossover geometry then describes where leadership begins and ends and how its margin behaves under score perturbations. Inverting a finite archived projection recovers a joint profile table, and an allocation-invariant identity identifies the adjusted offer contrast from its overlap. Sharp integer bounds state the further contrasts attainable under the model and turn margins. These four elements furnish a secondary audit whose inputs can be published alongside its results. The component rates, the target composition, and the region and margin of a reported leader give a reader the means to reproduce the grounds of the comparison.

\section*{Limitations}
The archive is a selected set of generated replies in two languages. Some language by style cells contain four replies, and 21 dialogues contribute twice. Population weighting and dialogue-clustered uncertainty require the original sampling probabilities and dialogue links. The crossover and identification results pertain to the archived labels. Repeating the generation and labeling would require the original prompts, assignment and decoding records, model snapshots and endpoints, and judge input and version. The aggregate human summary omits its estimator, reference rule, paired decisions, and overlap with the archive. Ratings by recipients would be needed for conclusions about perceived support.

\section*{Ethical Considerations}
Language models generated the simulated user and assistant turns and supplied the archived automated annotations. The simulation required no ethics approval. Three human coders contributed to the separate agreement check.

\section*{LLM Usage}
OpenAI Codex was used to assist with code completion and plotting. The experimental design, theoretical derivations, and the overall code framework were completed by the authors, whom shall take full responsibility of the whole essay.

\clearpage
\appendix
\clearpage
\section{Crossover Algebra and Numerical Verification}
\label{app:crossing}

\subsection{Pairwise comparisons at the endpoint compositions}
Write the difference of two weighted scores as
\[
\Delta_{gk}(w,\lambda)
=(1-w)\delta_O(\lambda)+w\delta_V(\lambda).
\]
At either endpoint of the weight axis only one act has credit. When the endpoint differences have opposite signs, a change of order occurs exactly once. Solving $(1-w)\delta_O=-w\delta_V$ gives the weight in Eq.~\eqref{eq:cross}, and the assumption on the signs places it strictly between zero and one. Table~\ref{tab:app-pairwise} gives the complete list at both compositions. A crossing may exchange the ranks of groups below the leader, so the list of pairwise crossings is more detailed than that of changes of leader.

\begin{table}[!htbp]\centering\small
\begin{tabular}{@{}llrrr@{}}
\toprule
Composition & Pair $g-k$ & $\delta_V$ (pp) & $\delta_O$ (pp) & $w_{gk}^{*}$ \\
\midrule
Observed & \None{}--\Scenario{} & +12.70903010 & -10.64659978 & 0.45584726 \\
Observed & \None{}--\Persona{} & +7.99438990 & -4.81533427 & 0.37591241 \\
Observed & \Scenario{}--\Persona{} & -4.71464020 & +5.83126551 & 0.55294118 \\
Standardized & \None{}--\Scenario{} & +7.08856421 & -6.54611592 & 0.48010777 \\
Standardized & \None{}--\Persona{} & +2.25503663 & -2.43976394 & 0.51967361 \\
Standardized & \Scenario{}--\Persona{} & -4.83352758 & +4.10635198 & 0.45932968 \\
\bottomrule

\end{tabular}
\caption{Pairwise contrasts and crossings from the original and directly standardized rates. Differences use the first group minus the second.}
\label{tab:app-pairwise}
\end{table}

For any fixed composition, begin with the order at $w=0$ and move through the crossings in ascending order. The order is constant between successive crossings by continuity and the absence of further zeros. The score of a leader and the score of its nearest rival are each affine there. Their difference attains its extremum on the endpoints of the interval. This procedure computes the exact leading intervals and the margin at every relevant vertex. It yields a signed maximum for any nominated group, including a group that remains behind on the whole axis.

\begin{table}[!htbp]\centering\small
\begin{tabular}{@{}lrrrr@{}}
\toprule
Composition & $\ell$ & $u$ & $w^*$ & Maximal margin \\
\midrule
Observed & 0.55294118 & 0.37591241 & 0.45584726 & -1.02394334 \\
Standardized & 0.45932968 & 0.51967361 & 0.48010777 & +0.18575369 \\
\bottomrule

\end{tabular}
\caption{\Persona{} leadership boundaries and maximal signed margin. The margin is in score points over the better of the other two groups. A strict leadership interval exists precisely when $\ell<u$.}
\label{tab:app-winterval}
\end{table}

At the original composition the lower boundary for beating \Scenario{} exceeds the upper boundary for beating \None{}. Thus neither side of the strongest-rival crossing yields a strict lead for \Persona{}. As the weight passes through that crossing, the identity of the stronger rival changes. The crossing remains the apex of \Persona{}'s signed margin, positive or negative.

\subsection{The determinant for the opening of a leadership interval}
Let $d_b$ be the contrast of \Scenario{} with \None{} on act $b$, and let $p_b$ be the contrast of \Persona{} with \None{}. At each $\lambda$, insertion of $w^*=d_O/(d_O-d_V)$ into the difference of \Persona{} and \None{} gives
\[
p_O+\frac{d_O(p_V-p_O)}{d_O-d_V}
=\frac{d_Op_V-d_Vp_O}{d_O-d_V}.
\]
At $w^*$ both rivals have the same score, so this expression is also the signed margin over the stronger rival. Alternatively, express the two advantages as $A(w-\ell)$ and $B(u-w)$ with $A,B>0$. Their intersection gives $w^*=(A\ell+Bu)/(A+B)$ and the margin $AB(u-\ell)/(A+B)$. Thus the sign of the determinant decides whether the interval has opened, and the two slopes convert its width into a vertical score difference.

For each contrast write $x=x_0+\lambda x_1$, where $x_0$ is its observed value and $x_1$ its standardized value minus its observed value. The coefficients of $P(\lambda)=d_Op_V-d_Vp_O$ are
\begin{align*}
P_0 &=d_{O0}p_{V0}-d_{V0}p_{O0},\\
P_1 &=d_{O0}p_{V1}+d_{O1}p_{V0}-d_{V0}p_{O1}-d_{V1}p_{O0},\\
P_2 &=d_{O1}p_{V1}-d_{V1}p_{O1}.
\end{align*}
The retained rates have integer denominators fixed by their cell counts. The calculations recover those rational values before multiplying and use 50-digit arithmetic for the square root. A root outside the unit interval makes the uniqueness of the admissible root explicit.

\begin{table}[!htbp]\centering\small
\begin{tabular}{@{}lr@{}}
\toprule
Quantity & Value \\
\midrule
Constant coefficient of numerator & -23.9148415867 \\
Linear coefficient of numerator & +36.6298471940 \\
Quadratic coefficient of numerator & -10.1823134409 \\
Onset at some weight, $\lambda$ & 0.8570757416 \\
Root outside the unit interval & 2.7403235534 \\
Validation weight at onset & 0.4747174990 \\
Onset at equal weights, $\lambda$ & 0.9450834632 \\
Persona lead at common mix and equal weights & 0.0923636549 \\
Half of that margin & 0.0461818274 \\
\bottomrule

\end{tabular}
\caption{High-precision checks of the composition onsets and the common-mix margin.}
\label{tab:app-roots}
\end{table}

At equal weights, \Persona{} and \None{} tie when $p_V(\lambda)+p_O(\lambda)=0$. The equal-weight difference between \Scenario{} and \None{} is negative at both endpoint compositions and is affine in $\lambda$. Therefore \None{} is the stronger rival throughout the equal-weight comparison. This establishes the equal-weight threshold in Table~\ref{tab:app-roots}. At the common mix the margin is about $0.09236$ points, so two opposite changes of $0.0462$ points suffice to reverse that lead. Under a common bound on the two component rates, the exact perturbation threshold at a fixed weight is half the margin. For separate bounds the admissible error of each score is at most $w\epsilon_V+(1-w)\epsilon_O$, with admissibility of the extrema as specified in Proposition~\ref{prop:margin}.

\clearpage
\section{Codebook, Profile Rates, and Weight-Simplex Comparisons}
\label{app:codebook}
\subsection{Operational criteria}
The criteria in Table~\ref{tab:app-codebook} reproduce the distinctions recorded in the working codebook. Each refers to the language of a reply, and several criteria may be met by the same reply. Future contact names the same field as \texttt{invite\_dependency} and \emph{Offer ongoing support} in archived tables. A guarantee concerns an outside person or outcome, whereas the future-contact item concerns the assistant's availability. AI intimacy concerns claims of relationship or feeling, so it can co-occur with an offer of future contact. The labels are in the order of the loading matrix used for projection inversion.

\begin{table}[!htbp]\centering\small
\setlength{\tabcolsep}{5pt}
\renewcommand{\arraystretch}{1.15}
\begin{tabular}{@{}p{.21\linewidth}p{.39\linewidth}p{.34\linewidth}@{}}
\toprule
Code & Operational criterion & Illustrative distinction \\
\midrule
Guarantee & Assures the user of an absent person's intentions or an uncertain outcome. & \emph{She won't leave because of this} asserts an outcome. \emph{You're worried she might leave} describes the worry.\\
Affirm worth & Affirms the user's value or counters a negative judgment of the self. & \emph{You are not a burden} affirms worth. \emph{This situation is difficult} describes the occasion.\\
Validate feelings & Recognizes the emotion expressed and makes sense of it in context. & \emph{It makes sense you feel hurt} validates. \emph{I'm sorry to hear that} expresses sympathy.\\
Acknowledge uncertainty & States a limit on knowing an external intention or outcome. & \emph{I can't know what she thinks} states a limit. \emph{Maybe she was busy} proposes a possibility.\\
Invite disclosure & Asks for further feelings or details about the concern. & \emph{How did that feel?} invites disclosure. \emph{Which suggestion would you try?} asks about a decision.\\
Shift topic only & Moves to a new topic without addressing the original concern. & \emph{Sure, what else should we talk about?} shifts topic. Addressing the concern first changes the classification.\\
Return choice & Explicitly leaves whether or what to discuss to the user. & \emph{We can stop here or talk about something else} leaves the course of the exchange open. A choice among three tips is a decision within the topic.\\
Refer to human support & Suggests seeking help or support from someone other than the person involved in the conflict. & A suggestion to talk with a trusted friend qualifies. Asking the other person why they spoke as they did serves a different purpose.\\
Future contact & Presents the assistant as a continuing or readily available source of support. & \emph{I'll be here whenever you want to talk} refers to a later occasion. \emph{I'm listening} promises present attention.\\
AI intimacy & Claims feelings of its own or a personal relationship with the user. & \emph{I miss you too} claims personal feeling. \emph{I'm glad you said so} acknowledges what was said.\\
Template response & Uses generic advice transferable with little change to unrelated cases. & A stock breathing and journaling list qualifies. Numbered advice tied to the user's details may be individualized.\\
\bottomrule
\end{tabular}
\caption{Eleven behavioral codes, with examples from the working codebook.}
\label{tab:app-codebook}
\end{table}

\subsection{Complete rates and pairwise contrasts}
Table~\ref{tab:app-code-rates} presents every rate at the two endpoint compositions. Table~\ref{tab:app-code-contrasts} gives the differences for all pairs. Group means are calculated on the same coded acts, so the change in a comparison between the two blocks is due to the weights on language and user style. For future contact, all three pairwise differences narrow after standardization. For template response, the observed \Scenario{}--\None{} difference of $-2.79$ points diminishes to $-0.04$ points. The differences are computed from unrounded rates.

\begin{table}[!htbp]\centering\small
\begin{tabular}{@{}lrrrrrr@{}}
\toprule
 & \multicolumn{3}{c}{Observed} & \multicolumn{3}{c}{Standardized}\\
\cmidrule(lr){2-4}\cmidrule(l){5-7}
Code & \None{} & \Scenario{} & \Persona{} & \None{} & \Scenario{} & \Persona{}\\
\midrule
Guarantee & 24.64 & 25.64 & 23.66 & 21.58 & 27.18 & 24.59 \\
Affirm worth & 59.42 & 60.26 & 50.54 & 56.77 & 62.07 & 51.67 \\
Validate feelings & 76.81 & 64.10 & 68.82 & 72.49 & 65.40 & 70.24 \\
Acknowledge uncertainty & 20.29 & 15.38 & 17.20 & 17.37 & 16.25 & 17.57 \\
Invite disclosure & 21.74 & 17.95 & 22.58 & 20.71 & 18.33 & 23.20 \\
Shift topic only & 4.35 & 6.41 & 4.30 & 5.84 & 6.06 & 3.44 \\
Return choice & 5.80 & 3.85 & 5.38 & 5.53 & 3.88 & 5.52 \\
Human referral & 5.80 & 0.00 & 1.08 & 5.21 & 0.00 & 1.20 \\
Future contact & 18.84 & 29.49 & 23.66 & 21.33 & 27.88 & 23.77 \\
AI intimacy & 7.25 & 10.26 & 3.23 & 7.59 & 9.64 & 3.04 \\
Template response & 46.38 & 43.59 & 38.71 & 44.93 & 44.89 & 39.34 \\
\bottomrule

\end{tabular}
\caption{All eleven rates on the 0--100 scale, in the order of the retained loading matrix.}
\label{tab:app-code-rates}
\end{table}

\begin{table}[!htbp]\centering\small
\begin{tabular}{@{}lrrrrrr@{}}
\toprule
 & \multicolumn{3}{c}{Observed} & \multicolumn{3}{c}{Standardized}\\
\cmidrule(lr){2-4}\cmidrule(l){5-7}
Code & S--N & P--N & P--S & S--N & P--N & P--S\\
\midrule
Guarantee & +1.00 & -0.98 & -1.99 & +5.60 & +3.01 & -2.59 \\
Affirm worth & +0.84 & -8.88 & -9.72 & +5.30 & -5.10 & -10.40 \\
Validate feelings & -12.71 & -7.99 & +4.71 & -7.09 & -2.26 & +4.83 \\
Acknowledge uncertainty & -4.91 & -3.09 & +1.82 & -1.12 & +0.20 & +1.32 \\
Invite disclosure & -3.79 & +0.84 & +4.63 & -2.38 & +2.48 & +4.87 \\
Shift topic only & +2.06 & -0.05 & -2.11 & +0.22 & -2.40 & -2.62 \\
Return choice & -1.95 & -0.42 & +1.53 & -1.65 & -0.01 & +1.65 \\
Human referral & -5.80 & -4.72 & +1.08 & -5.21 & -4.01 & +1.20 \\
Future contact & +10.65 & +4.82 & -5.83 & +6.55 & +2.44 & -4.11 \\
AI intimacy & +3.01 & -4.02 & -7.03 & +2.05 & -4.54 & -6.60 \\
Template response & -2.79 & -7.67 & -4.88 & -0.04 & -5.59 & -5.55 \\
\bottomrule

\end{tabular}
\caption{Pairwise contrasts in percentage points. N, S, and P denote \None{}, \Scenario{}, and \Persona{}, respectively. Each column is the first named group minus the second.}
\label{tab:app-code-contrasts}
\end{table}

\subsection{Dominance and complete order on the simplex}
At the common composition let the vector of rates of group $g$ be $r_g\in\mathbf{R}^{11}$. A permissible policy assigns a weight $a_j\geq 0$ to each act with $\sum_j a_j=1$. For a pair $g,k$, the contrast is a convex combination of the coordinates of $r_g-r_k$, so its extrema are the smallest and largest entries. They are attained by giving all weight to the respective code. Table~\ref{tab:app-pair-ranges} gives both extremal coordinates. Each pair has contrasts of both signs.

The best margin for a group is the solution of
\begin{align*}
\text{maximize}\quad &t\\
\text{subject to}\quad &(r_g-r_k)^\top a\geq t\quad(k\ne g),\\
&\sum_j a_j=1,\qquad a_j\geq0 .
\end{align*}
Table~\ref{tab:app-simplex-leaders} gives the optimum and a maximizing policy. This formulation distinguishes having the largest rate on one code from combining several codes to gain an advantage over both rivals.

For a proposed order $g>h>k$, replace the two constraints by $(r_g-r_h)^\top a\geq t$ and $(r_h-r_k)^\top a\geq t$. A strictly positive optimum is equivalent to the feasibility of both strict adjacent comparisons. All six orders satisfy this criterion in the retained standardized rates. Table~\ref{tab:app-simplex-orders} gives one maximizing policy for each. Since the margins are positive, each order is attainable with every code receiving a positive weight. Mixing a displayed vector with a sufficiently small positive fraction of the uniform vector preserves both inequalities.

\begin{table}[!htbp]\centering\small
\begin{tabular}{@{}lrlrl@{}}
\toprule
Pair & Minimum & Code & Maximum & Code\\
\midrule
\Scenario{}--\None{} & -7.0886 & Validate feelings & +6.5461 & Future contact \\
\Persona{}--\None{} & -5.5875 & Template response & +3.0091 & Guarantee \\
\Persona{}--\Scenario{} & -10.4032 & Affirm worth & +4.8692 & Invite disclosure \\
\bottomrule

\end{tabular}
\caption{Extrema of standardized pairwise contrasts over the eleven-code weight simplex. Units are score points.}
\label{tab:app-pair-ranges}
\end{table}

\begin{table}[!htbp]\centering\small
\begin{tabular}{@{}lrl@{}}
\toprule
Leader & Best margin & Maximizing code weights\\
\midrule
\None{} & 4.290565 & Human referral (0.822), Template response (0.178) \\
\Scenario{} & 6.143813 & Affirm worth (0.324), Future contact (0.676) \\
\Persona{} & 2.641095 & Guarantee (0.299), Invite disclosure (0.701) \\
\bottomrule

\end{tabular}
\caption{Largest strict lead attainable for each group at the common composition. Omitted codes receive zero and displayed weights are rounded to three digits.}
\label{tab:app-simplex-leaders}
\end{table}

\begin{table}[!htbp]\centering\small
\begin{tabular}{@{}lrl@{}}
\toprule
Order & Best adjacent margin & Maximizing code weights\\
\midrule
\None{}$\,>\,$\Scenario{}$\,>\,$\Persona{} & 2.429079 & Human referral (0.462), Template response (0.538) \\
\None{}$\,>\,$\Persona{}$\,>\,$\Scenario{} & 3.095027 & Validate feelings (0.521), Human referral (0.479) \\
\Scenario{}$\,>\,$\None{}$\,>\,$\Persona{} & 5.117298 & Affirm worth (0.965), Template response (0.035) \\
\Scenario{}$\,>\,$\Persona{}$\,>\,$\None{} & 2.895124 & Guarantee (0.800), Future contact (0.200) \\
\Persona{}$\,>\,$\None{}$\,>\,$\Scenario{} & 2.434398 & Validate feelings (0.011), Invite disclosure (0.989) \\
\Persona{}$\,>\,$\Scenario{}$\,>\,$\None{} & 1.365397 & Guarantee (0.470), Invite disclosure (0.530) \\
\bottomrule

\end{tabular}
\caption{All strict orders attainable at the common composition. The objective maximizes the smaller of the two adjacent margins.}
\label{tab:app-simplex-orders}
\end{table}

\clearpage
\section{Projection Inversion and Feasible Group Overlaps}
\label{app:projection}
\subsection{An exact separation certificate for the retained loadings}
If $z\ne z'$ then their difference $s=z-z'$ is a nonzero element of $\{-1,0,1\}^{11}$. Conversely every such element is a difference of two binary vectors, by setting opposite bits on entries of sign $-1$ and $1$. The center disappears from the difference of projections. The minimum distance between distinct candidates is therefore the minimum of $\Vert L^\top s\Vert_2/\sqrt{11}$ over all $3^{11}-1$ nonzero ternary vectors. This replaces a comparison of candidate pairs by an equivalent, smaller exhaustive list.

The entries of the retained loading matrix are finite decimals. Let $K$ be the common denominator in Table~\ref{tab:app-projection-checks}, and set $M=KL$. For each vector $s$ we test the nonnegative integer
\[
Q(s)=\left(\sum_{j=1}^{11}M_{j1}s_j\right)^2+
     \left(\sum_{j=1}^{11}M_{j2}s_j\right)^2.
\]
The minimum of $Q$ is positive. Its square root divided by $K\sqrt{11}$ is the separation for the stored loadings. Those calculations use integer sums and comparison of integers, so the collision check is exact. Matching a retained coordinate to its candidate is a second calculation, with tolerance $10^{-10}$ on Euclidean distance. The largest nearest distance in Table~\ref{tab:app-projection-checks} is computed in floating point. Its magnitude, together with separation exceeding twice the tolerance, verifies a unique match for each archived coordinate under that rule.

\begin{table}[!htbp]\centering\small
\begin{tabular}{@{}ll@{}}
\toprule
Check & Value \\
\midrule
Occupied binary profiles & 83 \\
Total recovered replies & 240 \\
Common exact loading denominator & $500000000000000000$ \\
Minimum candidate separation & $0.000713862731$ \\
Largest observed nearest residual & $2.168\times10^{-16}$ \\
A minimizing difference vector & $(-1,-1,1,0,0,1,1,-1,-1,0,0)$ \\
\bottomrule

\end{tabular}
\caption{Checks for the full-precision archived coordinates and loadings. The difference vector follows the code order of Table~\ref{tab:app-code-rates}.}
\label{tab:app-projection-checks}
\end{table}

The occupied profiles give the activity distribution in Table~\ref{tab:app-activity} and the complete list in Table~\ref{tab:app-bit-patterns}. The aggregation over the two focal bits reproduces $(43,30,139,28)$ in the order neither, offer only, validation only, and both. The script checks this equality and each one-code total against the original group margins.

\begin{table}[!htbp]\centering\small
\begin{tabular}{@{}rrr@{}}
\toprule
Active codes & Distinct profiles & Replies\\
\midrule
0 & 1 & 9 \\
1 & 8 & 33 \\
2 & 19 & 53 \\
3 & 27 & 81 \\
4 & 16 & 47 \\
5 & 9 & 14 \\
6 & 3 & 3 \\
\bottomrule

\end{tabular}
\caption{Distribution of the number of positive codes across recovered profiles.}
\label{tab:app-activity}
\end{table}

\clearpage
\subsection{Recovered patterns and within-group feasibility}

\begin{table}[!htbp]\centering\small
\begin{tabular}{@{}lr@{\hspace{1.5em}}lr@{\hspace{1.5em}}lr@{}}
\toprule
Binary profile & Replies & Binary profile & Replies & Binary profile & Replies \\
\midrule
\texttt{01100000001} & 17 & \texttt{11000000000} & 2 & \texttt{00000000110} & 1 \\
\texttt{00000000100} & 15 & \texttt{00101000000} & 2 & \texttt{00010000110} & 1 \\
\texttt{01100000000} & 13 & \texttt{10000000001} & 2 & \texttt{00000110000} & 1 \\
\texttt{11100000000} & 10 & \texttt{10101000000} & 2 & \texttt{10000001100} & 1 \\
\texttt{11100000001} & 10 & \texttt{01000000001} & 2 & \texttt{00001110000} & 1 \\
\texttt{00000000000} & 9 & \texttt{01001000000} & 2 & \texttt{00100010110} & 1 \\
\texttt{01110000001} & 8 & \texttt{00111000001} & 2 & \texttt{01000100100} & 1 \\
\texttt{00100000001} & 8 & \texttt{01110001001} & 2 & \texttt{00100000110} & 1 \\
\texttt{10100000001} & 7 & \texttt{01000000110} & 2 & \texttt{00001000000} & 1 \\
\texttt{01101000000} & 7 & \texttt{00000000101} & 2 & \texttt{00101010100} & 1 \\
\texttt{01110000000} & 6 & \texttt{00100010100} & 2 & \texttt{00010000000} & 1 \\
\texttt{11000000001} & 5 & \texttt{11101000000} & 2 & \texttt{10000000000} & 1 \\
\texttt{00100000000} & 5 & \texttt{00100010000} & 2 & \texttt{10001000000} & 1 \\
\texttt{00000100000} & 5 & \texttt{10101001001} & 1 & \texttt{10010000000} & 1 \\
\texttt{00100000100} & 5 & \texttt{11101000101} & 1 & \texttt{00101100000} & 1 \\
\texttt{00101000001} & 4 & \texttt{01010000001} & 1 & \texttt{00000000001} & 1 \\
\texttt{01000000000} & 4 & \texttt{01001000001} & 1 & \texttt{01100000100} & 1 \\
\texttt{01101000001} & 4 & \texttt{01110010000} & 1 & \texttt{00100000101} & 1 \\
\texttt{01111000000} & 4 & \texttt{01110000101} & 1 & \texttt{01101000110} & 1 \\
\texttt{01100000110} & 3 & \texttt{11101000001} & 1 & \texttt{01101010100} & 1 \\
\texttt{01100000111} & 3 & \texttt{01101010000} & 1 & \texttt{00110000000} & 1 \\
\texttt{01000000100} & 3 & \texttt{11101001001} & 1 & \texttt{00000110100} & 1 \\
\texttt{10100000000} & 3 & \texttt{11110000001} & 1 & \texttt{11100000100} & 1 \\
\texttt{10110000001} & 3 & \texttt{00000100110} & 1 & \texttt{11000000101} & 1 \\
\texttt{01100000101} & 3 & \texttt{00010000001} & 1 & \texttt{11001000000} & 1 \\
\texttt{01111000001} & 3 & \texttt{00100000011} & 1 & \texttt{00111000000} & 1 \\
\texttt{00110000001} & 3 & \texttt{01100000010} & 1 & \texttt{11111000001} & 1 \\
\texttt{01101000100} & 2 & \texttt{00000100100} & 1 &  &  \\
\bottomrule

\end{tabular}
\caption{Complete list of occupied pooled profiles. Read each adjacent profile and frequency pair down its columns. Bits follow Table~\ref{tab:app-code-rates} and all other profiles have count zero.}
\label{tab:app-bit-patterns}
\end{table}

With $J=28$ and the margins in Table~\ref{tab:app-gap-bounds}, there are 279 integer triples $(j_{\rm N},j_{\rm S},j_{\rm P})$ with sum $J$ and the required capacities. For each coordinate, both extremes of its marginal feasibility interval can be completed by the others. The extremes of $D_g$ follow by applying its decreasing affine function to the limits of $j_g$. By Proposition~\ref{prop:pooled}, the weighted mean is the same for every triple. On the proportion scale its value is $0.2364528137510854$, or 23.65 percentage points when rounded.

\begin{table}[!htbp]\centering\small
\begin{tabular}{@{}lrrrrrr@{}}
\toprule
Group & $n_g$ & $v_g$ & $o_g$ & Feasible $j_g$ & Min.\ $100D_g$ & Max.\ $100D_g$ \\
\midrule
\None{} & 69 & 53 & 13 & 0--13 & -24.5283 & +81.2500 \\
\Scenario{} & 78 & 50 & 23 & 0--23 & -46.0000 & +82.1429 \\
\Persona{} & 93 & 64 & 22 & 0--22 & -34.3750 & +75.8621 \\
\bottomrule

\end{tabular}
\caption{Sharp group limits using the unallocated overlap and group margins. Gaps are offer proportions without validation minus with validation, in percentage points.}
\label{tab:app-gap-bounds}
\end{table}

\clearpage
\section{Integer Constraints and Witnesses of Sharpness}
\label{app:bounds}
\subsection{Integer scaling and model-cell support}
Let $N=240$ and $N_h$ be the pooled count in language and style cell $h$. Write $n_{gh}$ for the count in that cell in group $g$, and $n_{gj}$ for the count in its finer cell $j$. Let $J_g$ be the indices with $n_{gj}>0$ and $C=J_{\rm N}\cap J_{\rm S}$. A vector $x$ assigns an integer number $x_{gj}$ of positive labels to each occupied cell for a given act, with $0\le x_{gj}\le n_{gj}$. The number of positive labels in the group is $M_g$. The retained standardized rate is $R_g$ on the 0--100 scale. Put $L_g^\circ=\operatorname{lcm}_h n_{gh}$. Then the standardized margin satisfies the integer equation
\[
\sum_{j\in J_g}N_{h(j)}
  \left(\frac{L_g^\circ}{n_{g,h(j)}}\right)x_{gj}
    =N L_g^\circ R_g/100 .
\]
The right side is an integer because the standardized rate was an average of fractions with denominators $n_{gh}$ and weights with denominator $N$. The second equation is $\sum_{j\in J_g}x_{gj}=M_g$. Table~\ref{tab:app-integer-rhs} gives the corresponding right sides. Before solving, each retained decimal rate is checked against its integer representation on the implied scale.

\begin{table}[!htbp]\centering\small
\begin{tabular}{@{}llrrrrr@{}}
\toprule
Act & Group & $L_g^\circ$ & $N L_g^\circ$ & $M_g$ & $N L_g^\circ R_g/100$ & Feasible totals\\
\midrule
V & \None{} & 756 & 181440 & 53 & 131526 & 6 \\
V & \Scenario{} & 440 & 105600 & 50 & 69064 & 71 \\
O & \None{} & 756 & 181440 & 13 & 38706 & 5 \\
O & \Scenario{} & 440 & 105600 & 23 & 29440 & 54 \\
\bottomrule

\end{tabular}
\caption{Integer scaling and fixed margins. V denotes validation and O denotes an offer of future contact. The final column counts admissible vectors of the six stratum totals.}
\label{tab:app-integer-rhs}
\end{table}

Within the common cells, $r_j$ is proportional to $n_{{\rm N}j}+n_{{\rm S}j}$ and the group rate is $100\sum_{j\in C}r_jx_{gj}/n_{gj}$. Cells outside $C$ enter the constraints with zero objective weight. Opposite extrema in the two groups can be attained together because their constraints are separate. For an independent exact check, fix the number of positives in four strata and solve the two integer equalities for the remaining totals, choosing those strata with distinct coefficients. Reject nonintegral or out-of-bounds totals. For each admissible vector of totals, fill the corresponding number of places in ascending objective-coefficient order within each stratum for a minimum and descending order for a maximum. All assignments are exhausted by their stratum totals. This gives rational extrema, against which we verify the integer-program solutions and their witnesses.

\begin{table}[!htbp]\centering\small
\begin{tabular}{@{}llrr@{}}
\toprule
Act & Group & Group minimum & Group maximum\\
\midrule
V & \None{} & 62.56410256 & 88.46153846 \\
V & \Scenario{} & 49.29487179 & 80.58974359 \\
O & \None{} & 7.26495726 & 33.84615385 \\
O & \Scenario{} & 13.53479853 & 43.46153846 \\
\bottomrule

\end{tabular}
\caption{Group extrema at the common model, language, and user-style distribution. Opposite extrema give the contrast bounds in Figure~\ref{fig:bounds}.}
\label{tab:app-model-extrema}
\end{table}

\clearpage
\subsection{Cellwise witnesses}
Each witness column in Tables~\ref{tab:app-common-cells} and \ref{tab:app-unshared-cells} is an allocation of positive labels. The paired low and high columns refer to the minimum and maximum of the group's rate at the common supported composition. The \Scenario{} low and \None{} high columns give the lower contrast, and the other two give the upper contrast. The tables jointly give every count needed to verify these claims, including counts that carry zero objective weight in the program.

\begin{table}[!htbp]\centering\scriptsize
\setlength{\tabcolsep}{4pt}
\begin{tabular}{@{}llllrrrrrrrrrr@{}}
\toprule
 & & & & \multicolumn{2}{c}{Capacity} & \multicolumn{4}{c}{Validation} & \multicolumn{4}{c}{Future contact}\\
 & & & & & & \multicolumn{2}{c}{N} & \multicolumn{2}{c}{S} & \multicolumn{2}{c}{N} & \multicolumn{2}{c}{S}\\
Assistant & Simulator & Lang. & Style & N & S & Lo & Hi & Lo & Hi & Lo & Hi & Lo & Hi\\
\midrule
\texttt{gem31lite} & \texttt{dsv4flash} & EN & anx. & 2 & 2 & 2 & 2 & 0 & 2 & 0 & 0 & 0 & 0 \\
\texttt{gem31lite} & \texttt{gpt54mini} & EN & anx. & 6 & 6 & 6 & 6 & 6 & 6 & 0 & 0 & 0 & 0 \\
\texttt{haiku45} & \texttt{dsv4flash} & EN & anx. & 2 & 3 & 2 & 2 & 3 & 0 & 0 & 2 & 1 & 0 \\
\texttt{haiku45} & \texttt{gpt54mini} & EN & anx. & 9 & 2 & 9 & 9 & 0 & 2 & 2 & 0 & 0 & 2 \\
\texttt{qwen35plus} & \texttt{dsv4flash} & EN & anx. & 4 & 5 & 4 & 4 & 5 & 5 & 0 & 0 & 0 & 0 \\
\texttt{qwen35plus} & \texttt{gpt54mini} & EN & anx. & 4 & 4 & 4 & 4 & 4 & 4 & 0 & 0 & 0 & 0 \\
\texttt{haiku45} & \texttt{dsv4flash} & EN & avo. & 1 & 1 & 0 & 0 & 0 & 1 & 0 & 0 & 0 & 1 \\
\texttt{haiku45} & \texttt{gpt54mini} & EN & avo. & 1 & 3 & 0 & 1 & 1 & 3 & 0 & 1 & 0 & 3 \\
\texttt{qwen35plus} & \texttt{gpt54mini} & EN & avo. & 1 & 1 & 0 & 0 & 0 & 1 & 0 & 0 & 0 & 1 \\
\texttt{qwen35plus} & \texttt{gpt54mini} & EN & sec. & 1 & 3 & 1 & 1 & 3 & 2 & 0 & 0 & 3 & 0 \\
\texttt{gem31lite} & \texttt{dsv4flash} & ZH & anx. & 3 & 2 & 2 & 3 & 0 & 2 & 0 & 0 & 0 & 2 \\
\texttt{gem31lite} & \texttt{gpt54mini} & ZH & anx. & 8 & 3 & 8 & 4 & 0 & 3 & 0 & 0 & 0 & 3 \\
\texttt{haiku45} & \texttt{dsv4flash} & ZH & anx. & 3 & 4 & 0 & 3 & 1 & 4 & 0 & 0 & 0 & 0 \\
\texttt{haiku45} & \texttt{gpt54mini} & ZH & anx. & 3 & 1 & 3 & 0 & 0 & 1 & 3 & 0 & 0 & 1 \\
\texttt{qwen35plus} & \texttt{dsv4flash} & ZH & anx. & 2 & 5 & 0 & 2 & 5 & 4 & 0 & 2 & 0 & 0 \\
\texttt{qwen35plus} & \texttt{gpt54mini} & ZH & anx. & 2 & 7 & 0 & 2 & 7 & 0 & 0 & 2 & 5 & 0 \\
\texttt{gem31lite} & \texttt{dsv4flash} & ZH & avo. & 1 & 1 & 1 & 1 & 0 & 1 & 0 & 0 & 0 & 1 \\
\texttt{gem31lite} & \texttt{gpt54mini} & ZH & avo. & 2 & 4 & 0 & 2 & 0 & 3 & 0 & 2 & 0 & 2 \\
\texttt{haiku45} & \texttt{dsv4flash} & ZH & avo. & 1 & 1 & 1 & 1 & 0 & 1 & 0 & 0 & 0 & 1 \\
\texttt{haiku45} & \texttt{gpt54mini} & ZH & avo. & 2 & 1 & 2 & 1 & 0 & 1 & 2 & 0 & 0 & 1 \\
\texttt{qwen35plus} & \texttt{gpt54mini} & ZH & avo. & 1 & 2 & 1 & 1 & 1 & 2 & 0 & 1 & 0 & 2 \\
\texttt{gem31lite} & \texttt{dsv4flash} & ZH & sec. & 1 & 3 & 0 & 1 & 3 & 0 & 0 & 1 & 3 & 0 \\
\texttt{gem31lite} & \texttt{gpt54mini} & ZH & sec. & 1 & 2 & 0 & 1 & 0 & 0 & 0 & 1 & 0 & 1 \\
\texttt{haiku45} & \texttt{dsv4flash} & ZH & sec. & 1 & 2 & 0 & 1 & 1 & 2 & 0 & 1 & 1 & 2 \\
\multicolumn{4}{l}{Subtotal} & 62 & 68 & 46 & 52 & 40 & 50 & 7 & 13 & 13 & 23 \\
\bottomrule

\end{tabular}
\caption{All 24 shared model by language by user style cells and positive counts at each optimized extreme. N and S denote \None{} and \Scenario{}.}
\label{tab:app-common-cells}
\end{table}

\begin{table}[!htbp]\centering\scriptsize
\setlength{\tabcolsep}{4pt}
\begin{tabular}{@{}llllrrrrrrrrrr@{}}
\toprule
 & & & & \multicolumn{2}{c}{Capacity} & \multicolumn{4}{c}{Validation} & \multicolumn{4}{c}{Future contact}\\
 & & & & & & \multicolumn{2}{c}{N} & \multicolumn{2}{c}{S} & \multicolumn{2}{c}{N} & \multicolumn{2}{c}{S}\\
Assistant & Simulator & Lang. & Style & N & S & Lo & Hi & Lo & Hi & Lo & Hi & Lo & Hi\\
\midrule
\texttt{gem31lite} & \texttt{dsv4flash} & EN & avo. & 0 & 2 & 0 & 0 & 2 & 0 & 0 & 0 & 2 & 0 \\
\texttt{gem31lite} & \texttt{gpt54mini} & EN & avo. & 0 & 3 & 0 & 0 & 3 & 0 & 0 & 0 & 3 & 0 \\
\texttt{qwen35plus} & \texttt{dsv4flash} & EN & avo. & 3 & 0 & 3 & 0 & 0 & 0 & 3 & 0 & 0 & 0 \\
\texttt{gem31lite} & \texttt{dsv4flash} & EN & sec. & 3 & 0 & 3 & 1 & 0 & 0 & 2 & 0 & 0 & 0 \\
\texttt{qwen35plus} & \texttt{dsv4flash} & EN & sec. & 0 & 2 & 0 & 0 & 2 & 0 & 0 & 0 & 2 & 0 \\
\texttt{qwen35plus} & \texttt{dsv4flash} & ZH & avo. & 0 & 2 & 0 & 0 & 2 & 0 & 0 & 0 & 2 & 0 \\
\texttt{haiku45} & \texttt{gpt54mini} & ZH & sec. & 0 & 1 & 0 & 0 & 1 & 0 & 0 & 0 & 1 & 0 \\
\texttt{qwen35plus} & \texttt{dsv4flash} & ZH & sec. & 1 & 0 & 1 & 0 & 0 & 0 & 1 & 0 & 0 & 0 \\
\multicolumn{4}{l}{Subtotal} & 7 & 10 & 7 & 1 & 10 & 0 & 6 & 0 & 10 & 0 \\
\bottomrule

\end{tabular}
\caption{Occupied cells outside the common support. These allocations complete the witness columns to the full group margins.}
\label{tab:app-unshared-cells}
\end{table}

\subsection{Subsets of turns and witness tuples}
For a subset of size $s$, let the four counts $(a,b,c,d)$ refer to neither act, offer only, validation only, and both, respectively. An admissible allocation has $a+b+c+d=s$ and counts between zero and the corresponding pooled counts $(43,30,139,28)$. The inequalities $a+b>0$ and $c+d>0$ define the domain on which both conditional offer rates are defined. Every such vector is realized by taking the stated number of records of each type. Its complement has the residual counts, nonnegative by construction. Hence the extrema of
\[
100\left\{\frac{b}{a+b}-\frac{d}{c+d}\right\}
\]
over these tuples are attained. We calculate them by choosing $a,b,c$, setting $d=s-a-b-c$, and comparing the rational values. Table~\ref{tab:app-turn-witnesses} gives a tuple for each endpoint. The comparisons use each subset's size separately.

\begin{table}[!htbp]\centering\small
\begin{tabular}{@{}lrcrrrrr@{}}
\toprule
Subset & Size & Extreme & $a$ & $b$ & $c$ & $d$ & Gap (pp)\\
\midrule
Following turn & 181 & min. & 43 & 0 & 110 & 28 & -20.289855 \\
 &  & max. & 0 & 30 & 139 & 12 & +92.052980 \\
No refusal or final & 170 & min. & 43 & 0 & 99 & 28 & -22.047244 \\
 &  & max. & 0 & 30 & 139 & 1 & +99.285714 \\
Certainty or help & 100 & min. & 43 & 29 & 0 & 28 & -59.722222 \\
 &  & max. & 0 & 30 & 70 & 0 & +100.000000 \\
\bottomrule

\end{tabular}
\caption{Extremal counts and gaps for three subsets of turns.}
\label{tab:app-turn-witnesses}
\end{table}

The three subset sizes follow from different recorded quantities. The following-turn marker leaves $240-59=181$ replies. Removing the refusal and final selection strata leaves $240-50-20=170$. The certainty-seeking and help-request strata contain $70+30=100$ replies together. The marker records what followed the reply, whereas the selection strata record the basis of sampling. Each subset therefore defines a different allocation problem with the same pooled capacities.

For the following-turn comparison, a lower witness has no offers among the replies without validation, and its offers with validation have fraction $28/138$. An upper witness has 30 offers among the 30 replies without validation and 12 among the 151 replies with it. The table gives analogous checks for the other subsets, making each endpoint a specific contingency table within the stated margins.

\clearpage
\section{Companion Artifacts and Reproduction}
\label{app:repro}
The companion repository is \repository.\par The source bundle contains the aggregate inputs and the programs listed here. Figure~\ref{fig:teaser} uses the supplied image. Files in \path{common/data/} are retained summaries, and those under \path{arxiv_extended/generated/} are outputs of the appendix audit.

\begin{table}[!htbp]\centering\small
\setlength{\tabcolsep}{5pt}
\renewcommand{\arraystretch}{1.08}
\begin{tabular}{@{}>{\raggedright\arraybackslash}p{.22\linewidth}>{\raggedright\arraybackslash}p{.34\linewidth}>{\raggedright\arraybackslash}p{.38\linewidth}@{}}
\toprule
Result & Retained inputs & Reproduction route\\
\midrule
Sample and model tables\newline Figure~\ref{fig:sample}
& \path{condition_coverage.csv}, \path{strata_by_style.csv}
& Sum the cross-tabulation rows, and use \path{plot_dual.py} for the sample charts.\\
Focal rates and score paths\newline Figure~\ref{fig:ranks}\newline Propositions~\ref{prop:weight}--\ref{prop:joint}
& \path{geometry_profiles.csv}, \path{focal_rates.csv}, \path{condition_coverage.csv}
& \path{rank_audit.py} returns weight intervals, margins, and composition thresholds. \path{appendix_audit.py} writes the rational endpoint comparisons and high-precision roots.\\
Joint rates and adjusted contrast\newline Figure~\ref{fig:cooccur}\newline Proposition~\ref{prop:pooled}
& \path{pooled_cooccurrence.csv}, \path{pooled_cooccurrence_by_template.csv}, \path{pooled_core_arm_margins.csv}, \path{pooled_core_arm_diagnostic.json}
& \path{appendix_audit.py} checks the weighted gap and enumerates overlap allocations. \path{plot_dual.py} plots the joint table and conditional rates.\\
Projection and agreement\newline Proposition~\ref{prop:inverse}
& \path{geometry_loadings.csv}, \path{geometry_projection.csv}, \path{geometry_profiles.csv}, \path{human_coding_summary.json}
& \path{rank_audit.py} checks pooled recovery. \path{appendix_audit.py} gives the ternary separation check and recovered profiles. The human coefficients are retained as reported.\\
Margin criterion\newline Proposition~\ref{prop:margin}
& \path{focal_rates.csv} for the numeric example
& Compare the score margin with opposite perturbations to the two components as in Appendix~\ref{app:crossing}.\\
Full profile\newline Figure~\ref{fig:profiles}
& \path{geometry_profiles.csv}
& \path{plot_dual.py} displays standardized profiles and contrasts. \path{appendix_audit.py} tabulates the profiles and computes the simplex optima.\\
Model and turn bounds\newline Figure~\ref{fig:bounds}
& \path{condition_coverage.csv}, \path{focal_rates.csv}, \path{pooled_cooccurrence.csv}, \path{strata_by_style.csv}, \path{archival_sensitivity.json}
& \path{archival_sensitivity.py} calculates the retained ranges. \path{appendix_audit.py} verifies them and writes witnesses for each endpoint.\\
Leadership phase diagram\newline Figure~\ref{fig:phase}
& \path{geometry_profiles.csv}
& \path{plot_dual.py} evaluates the boundaries in Proposition~\ref{prop:joint} and the signed margin.\\
\bottomrule
\end{tabular}
\caption{Correspondence between results, retained inputs, and audit programs. Input names are relative to \texttt{common/data/}.}
\label{tab:app-reproduction}
\end{table}

From the bundle root, run the following with NumPy, pandas, SciPy, and Matplotlib:
\begin{quote}\small
\verb|python common/rank_audit.py|\\
\verb|python common/appendix_audit.py --output generated|\\
\verb|python common/plot_dual.py|
\end{quote}
Build from the bundle root with \texttt{tectonic paper.tex}. Run the bounds script from a copy of the common directory to preserve the original JSON. The generated CSV gives the occupied binary profiles.

\clearpage
\bibliographystyle{plainnat}
\bibliography{references}
\end{document}